%% file: root.tex
\documentclass[letterpaper, 10 pt, conference]{ieeeconf}  % Comment this line out if you need a4paper

\IEEEoverridecommandlockouts                              % This command is only needed if 
\usepackage{graphics} % for pdf, bitmapped graphics files
\usepackage{booktabs}
\usepackage{graphicx}
\usepackage{multirow}
\usepackage{float}
\usepackage{mathrsfs}
\usepackage{colortbl}
\usepackage{color}
\usepackage[table]{xcolor} % 引入xcolor包，并使用table选项以支持colortbl功能
\usepackage{gradient-text}
\usepackage{multirow}
\usepackage{soul}
\newcommand\NickName{{\gradientRGB{SocialNav-Map}{15, 200, 200}{0, 0, 255}}\xspace}
\definecolor{mygray}{gray}{.88}
\usepackage{bbding}
\usepackage{caption}
\usepackage{pdfpages}
\usepackage{makecell}
\usepackage{amssymb}
\usepackage{xspace}
\usepackage{marvosym}
\usepackage{etoolbox}

\usepackage{amsmath}
\usepackage{algorithm}
\usepackage[noend]{algpseudocode}
\usepackage{graphicx}
\usepackage{stfloats}
\usepackage{subfigure}
\usepackage{caption}
\usepackage{setspace}
 
\usepackage{lipsum}
\usepackage{caption}
\usepackage{indentfirst} 
\usepackage{hyperref}
\hypersetup{hypertex=true,
colorlinks=true,
linkcolor=blue,
anchorcolor=blue,
citecolor=black}

\usepackage{amsmath} % assumes amsmath package installed
\definecolor{lightblue}{rgb}{0.8,0.85,1} % 浅蓝色

\definecolor{darkblue}{rgb}{0.414902, 0.561765, 0.982353}

\begin{document}
\title{\LARGE \bf
\NickName: Dynamic Mapping with Human Trajectory Prediction for Zero-Shot Social Navigation
}

\author{\textbf{Lingfeng Zhang$^{1,2,3}$, Erjia Xiao$^{4}$, Xiaoshuai Hao$^{3,\dag}$, Haoxiang Fu$^{5}$, Zeying Gong$^{4}$} \\ 
\textbf{Long Chen$^{3}$, Xiaojun Liang$^{2,\text{\Envelope}}$, Renjing Xu$^{4}$, Hangjun Ye$^{3}$, Wenbo Ding$^{1,\text{\Envelope}}$}
\thanks{$^\dag$ Project Leader.}
\thanks{$^\text{\Envelope}$ Corresponding Authors.}
\\[4pt]
% \vspace{4pt}
$^{1}$Shenzhen Tsinghua International Graduate School, Tsinghua Univeristy
\\
\vspace{2pt}
$^{2}$Peng Cheng Laboratory, $^{3}$Xiaomi EV \\
\vspace{2pt}
$^{4}$The Hong Kong University of Science and Technology (Guangzhou)
\\
\vspace{2pt}
$^{5}$National University of Singapore
\vspace{2pt}
\\
{\tt\small zlf25@mails.tsinghua.edu.cn, ding.wenbo@sz.tsinghua.edu.cn}
\vspace{-13pt}
}

\maketitle

% \let\thefootnote\relax\footnotetext{$^{*}$ Equal contribution.}
% \let\thefootnote\relax\footnotetext{$^{\dagger}$ Project leaders.}
% \let\thefootnote\relax\footnotetext{$^{\text{\Letter}}$ Corresponding author.}

%%%%%%%%%%%%%%%%%%%%%%%%%%%%%%%%%%%%%%%%%%%%%%%%%%%%%%%%%%%%%%%%%%%%%%%%%%%%%%%%
\vspace{-0.5cm}
\begin{abstract}
% Social navigation in dynamic environments densely populated 
Social navigation in densely populated dynamic environments poses a significant challenge for autonomous mobile robots, requiring advanced strategies for safe interaction.
Existing reinforcement learning (RL)-based methods require over 2000+ hours of extensive training and often struggle to generalize to unfamiliar environments without additional fine-tuning, limiting their practical application in real-world scenarios.
To address these limitations, we propose \textit{\textbf{\NickName}}, a novel zero-shot social navigation framework that combines dynamic human trajectory prediction with occupancy mapping, enabling safe and efficient navigation without the need for environment-specific training.
Specifically, \textit{\textbf{\NickName}}  first transforms the task goal position into the constructed map coordinate system. Subsequently, it creates a dynamic occupancy map that incorporates predicted human movements as dynamic obstacles.
The framework employs two complementary methods for human trajectory prediction: history prediction and orientation prediction. By integrating these predicted trajectories into the occupancy map, the robot can proactively avoid potential collisions with humans while efficiently navigating to its destination.
Extensive experiments on the Social-HM3D and Social-MP3D datasets demonstrate that \textit{\textbf{\NickName}} significantly outperforms state-of-the-art (SOTA) RL-based methods, which require \textit{2,396 GPU hours} of training.
Notably, it reduces human collision rates by over 10\% without necessitating any training in novel environments. 
By eliminating the need for environment-specific training, \textit{\textbf{\NickName}} achieves superior navigation performance, paving the way for the deployment of social navigation systems in real-world environments characterized by diverse human behaviors.
The code is available at: https://github.com/linglingxiansen/SocialNav-Map.

\end{abstract}

% \vspace{-0.2cm}
%%%%%%%%%%%%%%%%%%%%%%%%%%%%%%%%%%%%%%%%%%%%%%%%%%%%%%%%%%%%%%%%%%%%%%%%%%%%%%%%
\section{INTRODUCTION}
% \vspace{-0.2cm}

\begin{figure}[!htbp]
% \vspace{-1cm}
\centering
\includegraphics[width=0.45\textwidth]{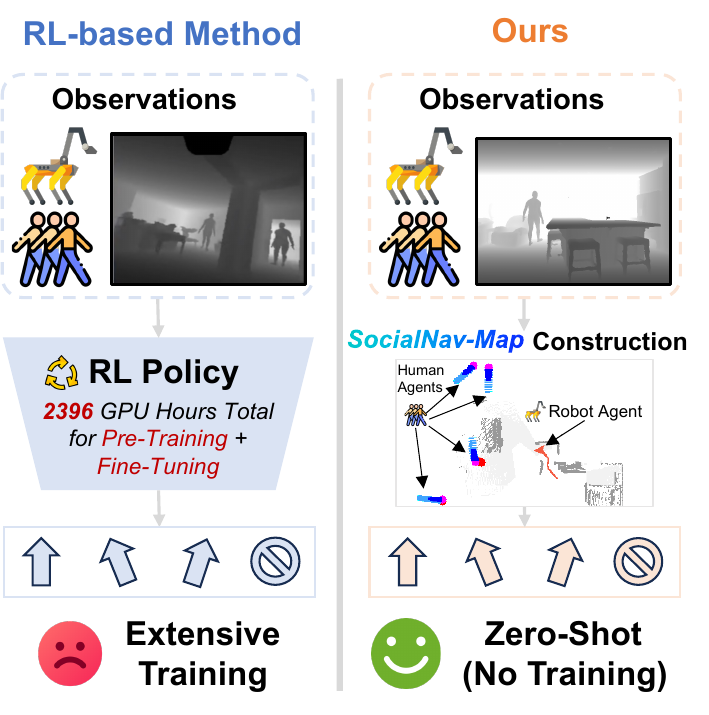}
\caption{\textbf{Comparison of reinforcement learning-based social navigation methods and our \textit{\NickName} framework.} 
Traditional reinforcement learning (RL)-based methods necessitate extensive training—totaling 2,396 GPU hours—through trial-and-error in simulated environments prior to deployment. In contrast, we present \textbf{\textit{\NickName}}, a zero-shot social navigation framework that tackles social navigation tasks by constructing dynamic occupancy maps and predicting human trajectories. Notably, \textbf{\textit{\NickName}} achieves superior performance compared to state-of-the-art RL-based methods, all without the need for any training.}
%Traditional reinforcement learning (RL)-based methods require extensive training (\textit{2,396 GPU hours total}) through trial-and-error in simulated environments before deployment. We propose \textbf{\textit{\NickName}}, a zero-shot social navigation framework. \textbf{\textit{\NickName}} accomplishes social navigation tasks by constructing dynamic occupancy maps and predicting human trajectories, while outperforming SOTA RL-based methods without any training.}
 \vspace{-0.5cm}
\label{fig1}
\end{figure}

Autonomous robots in densely populated environments face significant social navigation challenges, needing to achieve navigation goals while respecting social norms and ensuring human comfort~\cite{mavrogiannis2023core,perez2021robot,francis2023principles,xiao2025team}. This issue is especially critical indoors, where limited space, unpredictable human behavior, and complex layouts create dynamic obstacles that traditional navigation systems struggle to manage effectively~\cite{zhang2024trihelper,zhang2025multi,zhang2025novel,zhang2025humanoidpano,liu2025toponav,hao2022listen,hao2021matters,zhang2025vtla}.

The current approach to Social Navigation (SocialNav) primarily utilizes reinforcement learning (RL)~\cite{kapoor2023socnavgym,wang2024multi,hirose2024selfi,schulman2017proximal,zhang2025lips}, wherein agents develop navigation policies through extensive interactions within simulated environments. Despite showing potential in controlled settings, RL methods encounter significant limitations that impede practical deployment. They typically require extensive training periods—often several weeks—to formulate effective policies for specific environments, and these policies often struggle to generalize to new contexts with different layouts, human densities, or behaviors, necessitating costly retraining~\cite{gong2025cognition,schulman2017proximal,wijmans2019dd,team2025robobrain,wu2025evaluating}. This generalization issue is worsened by the reactive nature of most RL systems, which tend to prioritize immediate obstacle avoidance over long-term strategic planning. In complex scenarios, such as busy intersections or heavily trafficked corridors, these short-sighted strategies can result in inefficient navigation, deadlocks, or socially inappropriate behaviors that may discomfort those nearby~\cite{cancelli2023exploiting,nishimura2020risk}. Recent advancements in human trajectory prediction offer a promising alternative, allowing robots to move from passive to active navigation strategies by forecasting future human movements, thereby enhancing social acceptance and navigation efficiency. However, trajectory prediction in indoor environments presents challenges, including limited prediction ranges due to frequent directional changes, complex layouts that restrict movement, and high interaction densities that require precise collision avoidance~\cite{rudenko2020human,huang2023multimodal,fang2024spiking,zhang2025video,tang2025roboafford}.

As shown in Fig.~\ref{fig1}, researchers recently proposed Falcon~\cite{gong2025cognition}, a future-oriented framework that integrates explicit trajectory prediction into reinforcement learning-based SocialNav. Their approach effectively demonstrates the advantages of incorporating human trajectory prediction through auxiliary tasks during training, achieving state-of-the-art performance on the SocialNav benchmark. However, Falcon still requires extensive reinforcement learning training—\textbf{\textit{totaling 2,396 GPU hours for pre-training (2,204 hours)~\cite{ramakrishnan2021habitat} and fine-tuning (192 hours)~\cite{gong2025cognition}}}—and faces generalization challenges when deployed in new environments, necessitating additional fine-tuning for optimal performance.

% As shown in Fig.~\ref{fig1}, Gong et al. [20] recently proposed Falcon, a future-oriented framework that integrates explicit trajectory prediction into reinforcement learning-based social navigation. Their approach demonstrated the effectiveness of incorporating human trajectory prediction through auxiliary tasks during training and achieved state-of-the-art performance on the social navigation benchmark. However, Falcon still requires extensive RL training (\textit{\textbf{2496 GPU hours total}}) and faces generalization challenges when deployed to new environments without additional fine-tuning.

% To address this challenge, we present \textit{\textbf{\NickName}} as shown in Fig.~\ref{fig1}, a zero-shot social navigation framework that eliminates the need for environment-specific training while outperforming state-of-the-art reinforcement learning methods. Our framework first transforms the task goal location into a constructed map coordinate system and then constructs a socially aware dynamic occupancy map that incorporates predicted human motion trajectories as dynamic obstacles through a structured process.
To address this challenge, we present \textit{\textbf{\NickName}}, as shown in Fig.~\ref{fig1}, a zero-shot SocialNav framework that eliminates the need for environment-specific training while outperforming state-of-the-art reinforcement learning methods~\cite{zhang2024trihelper,zhang2025multi,gong2025stairway,zhang2025novel,zhang2025nava}. Our framework begins by transforming the task goal location into a constructed map coordinate system. It then creates a socially aware dynamic occupancy map that integrates predicted human motion trajectories as dynamic obstacles, employing a structured process to enhance navigation efficiency and safety.
Specifically, the core innovation of our approach lies in a fused trajectory prediction system that integrates two complementary methods: \textit{\textbf{historical prediction}}, which analyzes patterns derived from observed human motion, and \textit{\textbf{orientation prediction}}, which infers future trajectories based on current human pose and orientation. Utilizing these predictions, we treat both past and anticipated human trajectories as dynamic obstacles within the occupancy map, enabling proactive navigation decisions rather than merely reactive obstacle avoidance.
Past and predicted human positions are incorporated into the occupancy map as dynamic obstacles that are gradually updated over time. This design ensures that navigation decisions account for potential future human positions while preventing outdated predictions from creating persistent obstacles, which could lead to overly cautious or inefficient robot behavior.
Extensive experiments demonstrate that our zero-shot SocialNav framework can effectively complete tasks in dynamic social scenarios without any prior training, achieving state-of-the-art performance comparable to reinforcement learning methods that require significant training (\textbf{\textit{totaling 2,396 GPU hours}}). This underscores the performance and generalization capabilities of our approach.

% Specifically, the core innovation of our approach lies in a fused trajectory prediction system that combines two complementary approaches: \textit{\textbf{historical prediction}} (analyzing patterns learned from observed human motion) and \textit{\textbf{orientation prediction}} (inferring future trajectories based on current human pose and orientation). Based on these predictions, we treat past and predicted human trajectories as dynamic obstacles in the occupancy map, enabling proactive navigation decisions rather than passive obstacle avoidance. Past and predicted human positions are added to the occupancy map as dynamic obstacles and are gradually cleaned up over time. This design ensures that navigation decisions incorporate possible future human positions while preventing outdated predictions from causing persistent obstacles, which could lead to overly conservative or inefficient robot behavior.
% Extensive experiments demonstrate that our zero-shot social navigation framework can directly complete tasks in dynamic social scenarios without training, while achieving SOTA performance comparable to RL-based methods that require extensive training (\textit{\textbf{2496 GPU hours total}}), demonstrating the performance and generalization of our method. 

\indent The main contributions of our paper are as follows:

\begin{itemize}
\item We propose \textit{\textbf{\NickName}}, a zero-shot social navigation framework capable of effectively solving social navigation tasks in dynamic environments without the need for environment-specific training.

%We propose \textit{\textbf{\NickName}}, a zero-shot social navigation framework that can solve social navigation tasks in dynamic environments without requiring specific training.

\item 
We develop two complementary zero-shot human trajectory prediction methods: \emph{historical motion pattern analysis} and \emph{orientation-based linear extrapolation}, both of which effectively forecast future human trajectories.
%We develop two complementary zero-shot human trajectory prediction methods: historical motion pattern analysis and orientation-based linear extrapolation, which effectively predict future human trajectories.

\item 
%We introduce a dynamic obstacle management system that considers both current and predicted human positions as dynamic obstacles, utilizing an adaptive decay mechanism to facilitate efficient obstacle avoidance and navigation.
We introduce a dynamic obstacle management system that treats both the current and predicted human positions as dynamic obstacles and employs an adaptive decay mechanism to achieve efficient obstacle avoidance and navigation.

\item  
Extensive experiments demonstrate that our \textit{\textbf{\NickName}} achieves SOTA performance while reducing human collision rates by over 10\% without any training, and can be generalized to any scenario.
\end{itemize}

\section{RELATED WORK}

\subsection{Social Navigation}

Social Navigation (SocialNav)~\cite{perez2021robot,francis2023principles,xiao2025team, xia2020interactive} is a challenging robotics task that requires autonomous agents to navigate safely and efficiently in an environment where humans are moving and reach a specified target location. The main goal is to enable robots to understand, predict, and appropriately respond to human behavior, ensuring collision-free navigation while respecting social norms and personal space.
Various methods have been developed to address the challenges of social navigation. Reinforcement learning-based methods~\cite{team2025robobrain,zhang2025lips,gong2025cognition, cancelli2023exploiting} train agents through extensive interactions with a simulated environment and learn collision avoidance behaviors through trial and error. Multi-agent coordination methods~\cite{socialforces, socialattention} model the navigation problem as a distributed planning task where multiple agents coordinate their motion. Graph-based methods~\cite{socialforces, socialattention,social-graph,liu2025toponav} represent human-robot interactions through spatiotemporal graphs to capture agent dynamics and predict future motion over time. Recent studies have also explored egocentric navigation in photorealistic environments~\cite{ martin2021jrdb, vuong2023habicrowd,zhang2025nava,zhang2025team}, focusing on first-person perception and decision-making. Falcon helps agents perform safe social navigation by predicting human trajectories as auxiliary information. However, these methods typically require thousands of hours of training and struggle to generalize to new environments.
Unlike existing methods that rely on extensive environment-specific training, our approach introduces a zero-shot framework that combines occupancy mapping, trajectory prediction, and dynamic human obstacle mapping, enabling effective social navigation without requiring pre-training in the target environment.

\subsection{Zero-Shot Embodied Navigation}

Zero-Shot Embodied Navigation (ZSEN) has become a promising paradigm, addressing the limitations of learning-based approaches: these methods require extensive training, suffer from poor generalization, and struggle to adapt to new environments~\cite{wu2025evaluating,team2025robobrain,li2025vquala,jiang2024recursive}. For Object Navigation (ObjectNav), methods like CoW~\cite{gadre2023cows} and ESC~\cite{zhou2023esc} use CLIP~\cite{li2022grounded} for exploration and object detection. L3MVN~\cite{yu2023l3mvn}, TriHelper~\cite{zhang2024trihelper}, MFNP~\cite{zhang2025multi}, and VLFM~\cite{yokoyama2023vlfm} use semantic value map-based approaches for exploration and task completion. For Vision-and-Language Navigation (VLN), methods like NavGPT~\cite{zhou2024navgpt}, DiscussNav~\cite{long2024discuss}, and NavCoT~\cite{lin2025navcot} use large models for inference and action generation. Navid~\cite{zhang2024navid}, MapNav~\cite{zhang2025mapnav}, and NaVILA~\cite{cheng2024navila} use Vision-Language Models (VLMs) as a backbone for end-to-end training and deployment. InstrucNav~\cite{long2024instructnav} and UniGoal~\cite{yin2025unigoal} can simultaneously address both ObjectNav and VLN tasks. However, all previous methods are unable to directly address SocialNav task and handle dynamic human obstacles. We propose a zero-shot SocialNav framework and dynamic human obstacle occupancy map construction for the first time.

\begin{figure*}[!htbp]
% \vspace{-1cm}
\centering
\includegraphics[width=0.90\linewidth]{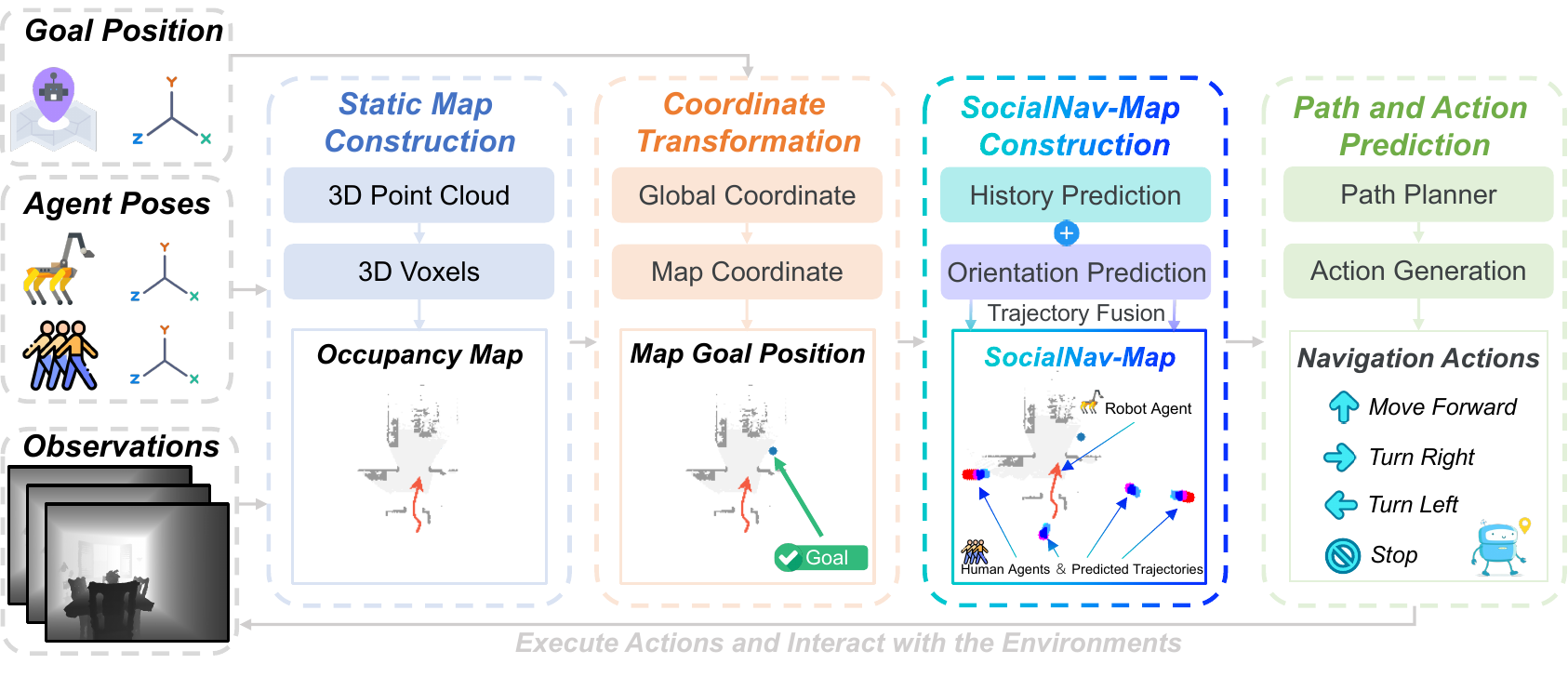}
\caption{\textbf{The general pipeline of our \textit{\NickName} framework.}  The system first constructs a static occupancy map from observation data through 3D point cloud generation and voxel discretization, then converts target locations into map coordinates. The core of \textbf{\textit{\NickName}} employs dual trajectory prediction—history prediction analyzes past human motion patterns, while direction prediction infers future positions—and fuses these predictions with dynamic obstacles with time decay. Finally, a fast marching method generates optimal collision-free paths and converts them into discrete navigation actions for zero-shot social navigation.}
\vspace{-0.5cm}
\label{fig2}
\end{figure*}

\label{sec:frontier}
\section{METHODOLOGY}

\subsection{Problem Formulation} 

The Social Navigation (SocialNav) task requires agents to navigate safely and efficiently to designated goal locations in environments populated by dynamic humans. The task is defined within scene $S$ containing $N$ moving humans $\mathcal{H} = \{h_1, h_2, ..., h_N\}$, where each human $h_i$ follows a trajectory $\tau_i(t)$ with position and orientation evolving over time. At the beginning of each episode, the agent is randomly initialized at position $p_0$ within scene $S$ and is assigned a specific goal location $g \in \mathcal{G}$.

At each discrete time step t, the agent receives an observation vector $\mathcal{O}_t = (V_t, P_t, H_t)$, where $V_t$ represents visual input (depth images), $P_t$ represents the agent's current pose, and $H_t$ represents observable human positions and orientations. Based on these inputs, the agent must select an action $a_t \in \mathcal{A}$, where the action space $\mathcal{A}$ contains four actions: move forward, turn left, turn right, and stop. The agent must navigate toward its goal while maintaining safe distances from both static obstacles and dynamic humans.

Task success is defined by two criteria: (1) the agent reaches within distance threshold $d_g$ of the goal location (typically 0.2 meters), and (2) the agent avoids collisions with humans throughout the episode, defined as maintaining minimum distance $d_h$ from any human (typically 0.1 meters). Each navigation episode is limited to $T_{max}$ time steps (typically 500 steps).

\subsection{Overview}
Our \textit{\textbf{\NickName}} framework operates through a four-stage pipeline that transforms observations $\mathcal{O}_t = (V_t, P_t, H_t)$ into socially-aware navigation actions. The Static Map Construction stage builds an occupancy map from RGB-D observations, while Coordinate Transformation converts the goal position $g$ into map coordinates. The core \textit{\textbf{\NickName}} Construction employs dual trajectory prediction: \textit{history prediction} analyzes human motion patterns and \textit{orientation prediction} extrapolates future positions based on current poses. These predicted trajectories are fused as dynamic obstacles with temporal decay into the occupancy map. Finally, Path and Action Prediction generates collision-free paths and converts them into discrete actions $a_t \in \mathcal{A}$ to achieve safe navigation without environment-specific training.

\subsection{Static Map Construction}
% The construction of the semantic map $\mathcal{M}$ employs the method proposed by \cite{chaplot2020object}, utilizing RGB-D images $V_t$ and the agent's ground truth pose data $P_t$. The map is populated by converting visual data into point clouds using geometric algorithms, which are then projected onto a 2D top-down view. This approach incorporates physical obstacles, explored areas, and semantically segmented object categories. Precise alignment between semantic masks and point clouds enables accurate channel mapping within the semantic map. The map is represented as a three-dimensional tensor with dimensions $C \times W \times H$, where $W$ and $H$ denote the map's width and length, respectively, and $C$ equals $C_n + 5$, with $n$ representing the number of object categories. The tensor's initial four channels encode obstacle information, explored terrain, current agent position, and historical agent locations. The subsequent $n$ channels delineate semantic maps for $n$ distinct object types, followed by an additional channel dedicated to stair mapping for subsequent analysis of stair navigation policy. At the beginning of each episode, the semantic map is initialized, with the agent's starting position defaulting to the map's central coordinates. Semantic maps serve as the foundational element that enables our system effectively navigate to target objects without requiring any prior training on the specific instances.

The static map construction process transforms RGB-D observations $V_t$ into a 2D occupancy map through a series of geometric transformations. Given a depth image $D \in \mathbb{R}^{H \times W}$ where $H$ and $W$ are image dimensions, we first generate a 3D point cloud using the camera intrinsic matrix $K$:
\begin{equation}
    P_{cam} = K^{-1} \begin{bmatrix} u \cdot D(u,v) \\ v \cdot D(u,v) \\ D(u,v) \end{bmatrix},
\end{equation}

where $(u,v)$ are pixel coordinates and $D(u,v)$ is the depth value at pixel $(u,v)$. The camera coordinate points $P_{cam}$ are then transformed to the agent's coordinate frame using the camera height $h_{agent}$ and elevation angle $\theta_{eve}$:
\begin{equation}
    P_{agent} = R_x(\theta_{eve}) \cdot P_{cam} + \begin{bmatrix} 0 \\ 0 \\ h_{agent} \end{bmatrix},
\end{equation}

where $R_x(\theta_{eve})$ is the rotation matrix around the x-axis. Next, we apply pose transformation to center the agent's view using shift location $s = [\frac{r \cdot res}{2}, 0, \frac{\pi}{2}]$ where $r$ is vision range and $res$ is resolution:
\begin{equation}
    P_{centered} = R_z(s_2) \cdot P_{agent} + \begin{bmatrix} s_0 \\ s_1 \\ 0 \end{bmatrix}.
\end{equation}
The centered 3D points are then discretized into voxel coordinates by normalizing to the range $[-1, 1]$:
\begin{equation}
    P_{voxel} = \frac{2 \cdot (P_{centered} - center)}{range} - 1,
\end{equation}
where center and range define the voxel grid bounds. Finally, the 3D voxel representation is projected to a 2D occupancy map by summing over height layers within the agent's traversable range $[z_{min}, z_{max}]$:
\begin{equation}
    M_{occ}(x,y) = \sum_{z=z_{min}}^{z_{max}} V_{voxel}(x,y,z).
\end{equation}

The resulting occupancy map $M_{occ}$ represents obstacles as high-confidence regions where depth measurements indicate solid surfaces within the agent's navigation height range.

% \vspace{-10pt}
\subsection{Coordinate Transformation}

The coordinate transformation stage establishes a consistent spatial reference by converting goal positions from world coordinates to map coordinates and maintaining fixed spatial relationships throughout navigation. Given the goal position in world coordinates $g_{world} = [x_w, y_w, z_w]$ and the agent's initial position $p_{init} = [x_{init}, y_{init}, z_{init}]$, the system first transforms world coordinates to simulator coordinates using:
\begin{equation}
    g_{sim} = \begin{bmatrix} -z_w \\ -x_w \\ y_w \end{bmatrix}, \quad p_{sim} = \begin{bmatrix} -z_{init} \\ -x_{init} \\ y_{init} \end{bmatrix}.
\end{equation}

The relative offset between goal and agent in simulator coordinates is computed as:
\begin{equation}
    \Delta_{sim} = g_{sim} - p_{sim} = \begin{bmatrix} -(z_w - z_{init}) \\ -(x_w - x_{init}) \\ y_w - y_{init} \end{bmatrix}.
\end{equation}

This offset is then mapped to the global map coordinate system, where the agent's initial position is fixed at the map center $p_{map\_center} = [\frac{W_{map}}{2}, \frac{H_{map}}{2}]$ with $W_{map}$ and $H_{map}$ representing map dimensions in meters. The goal's global map position is calculated as:
\begin{equation}
    g_{global\_map} = p_{map\_center} + \begin{bmatrix} -\Delta_{sim}[0] \\ \Delta_{sim}[1] \end{bmatrix}.
\end{equation}

For dynamic trajectory prediction, human positions $h_{world} = [x_h, y_h, z_h]$ undergo the same transformation pipeline using the fixed initial agent reference frame:
\begin{equation}
    h_{global\_map} = p_{map\_center} + \begin{bmatrix} -(-z_h + z_{init}) \\ -(-x_h + x_{init}) \end{bmatrix}.
\end{equation}

When the transformed goal position falls outside the current local map boundaries, the system places an intermediate goal in the direction of the true goal to maintain navigation progress. The system calculates the direction vector from the agent's current position to the out-of-bounds goal and places an intermediate target at distance $d_{\text{intermediate}} = \min(W_{\text{local}}/3, H_{\text{local}}/3)$  along this direction:
\begin{equation}
    g_{\text{intermediate}} = {a}_{\text{local}} + \frac{g_{\text{local}} - a_{\text{local}}}{||{g}_{\text{local}} - {a}_{\text{local}}||} \times d_{\text{intermediate}},
\end{equation}
where coordinates are clamped to remain within safe map boundaries $[3, map_{size}-4]$.

This coordinate transformation ensures spatial consistency by maintaining a fixed reference frame anchored to the agent's initial position, enabling accurate goal localization and human trajectory prediction regardless of the agent's current location during navigation.

\begin{figure*}[!htbp]
% \vspace{-1cm}
\centering
\includegraphics[width=0.88\linewidth]{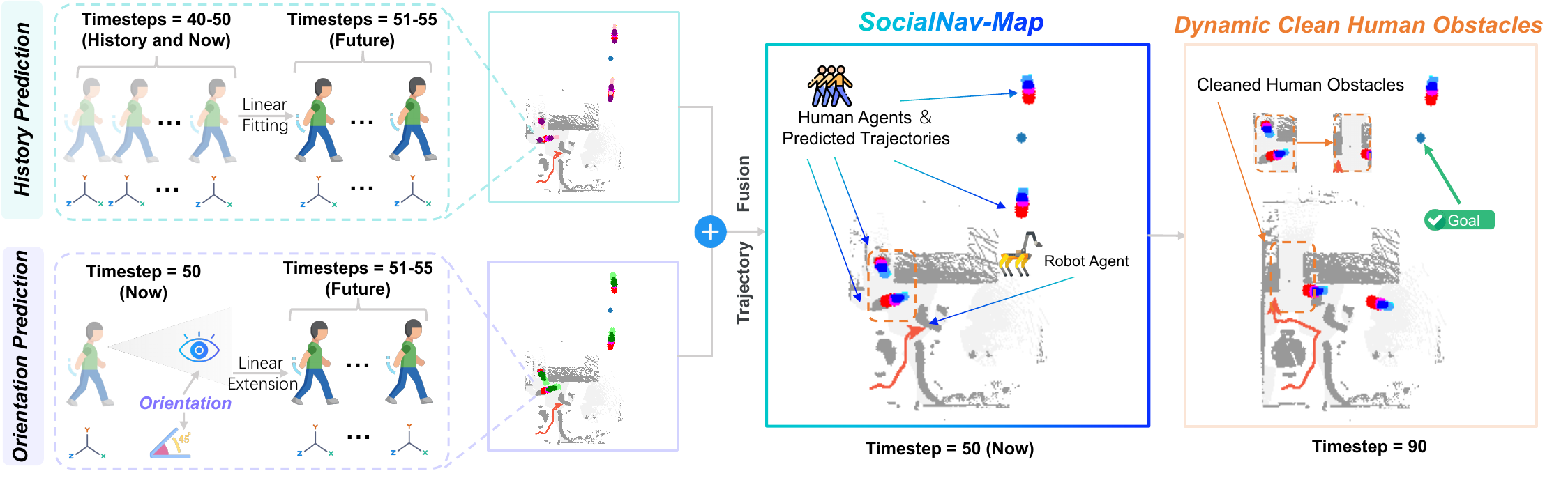}
\caption{\textbf{Human trajectory prediction and dynamic obstacle management in \textit{\NickName}.}  History prediction analyzes past human motion and uses linear regression fitting to infer future trajectories. Orientation prediction uses the current human pose and orientation to predict future positions via linear extension. These two prediction methods are fused to generate comprehensive trajectory predictions, which are integrated into the occupancy map as dynamic obstacles. These predicted human obstacles are automatically removed from the map after a certain decay period, preventing outdated predictions from hindering robot navigation while ensuring safety during active human movement.}
\vspace{-0.5cm}
\label{fig3}
\end{figure*}

\subsection{\textbf{\textit{\NickName}} Construction}
The general pipeline of our \textbf{\textit{\NickName}} Construction is illustrated in Fig.~\ref{fig2} and Fig.~\ref{fig3}. 
The \textbf{\textit{\NickName}} construction integrates dual trajectory prediction methods to anticipate human movements. The system maintains historical position data $H_i = \{(p_t, t)\}$ for each human agent $i$, where $p_t = [x_t, y_t]$ represents position at timestep $t$.

\textbf{History Prediction} For agents with sufficient trajectory history (length $\geq$ 2), the system performs linear regression on historical positions. Given position sequence $\{p_{t-n}, ..., p_{t-1}, p_t\}$ with corresponding timesteps $\{t-n, ..., t-1, t\}$, the future trajectory is predicted using:
\begin{equation}
    \hat{p}^{hist}_{t+k} = \begin{bmatrix} a_x \cdot (t+k) + b_x \\ a_y \cdot (t+k) + b_y \end{bmatrix},
\end{equation}

where $(a_x, b_x)$ and $(a_y, b_y)$ are linear coefficients obtained via least squares fitting over the historical trajectory data.

\textbf{Orientation Prediction} For each human with current orientation $\theta_i$, the system predicts future positions along the heading direction:
\begin{equation}
    \hat{p}^{orient}_{t+k} = p_t + k \cdot d_{step} \cdot \begin{bmatrix} \cos(\theta_i + \pi/2) \\ -\sin(\theta_i + \pi/2) \end{bmatrix},
\end{equation}

where $d_{step}$ is the step distance and the angle adjustment accounts for coordinate system transformation.

\textbf{Trajectory Fusion} The two prediction methods are fused using adaptive weighting that favors orientation prediction for longer horizons. The fused trajectory is computed as:
\begin{equation}
    \hat{p}^{fused}_{t+k} = w^{hist}_k \cdot \hat{p}^{hist}_{t+k} + w^{orient}_k \cdot \hat{p}^{orient}_{t+k},
\end{equation}

where $w^{hist}_k = w_{base} \cdot \alpha^k$ represents diminishing historical influence with decay factor $\alpha$, and $w^{orient}_k = 1 - w^{hist}_k$ ensures complementary weighting. This fusion approach leverages historical patterns for near-term predictions while relying on orientation-based extrapolation for longer-term forecasting, providing robust trajectory estimation across varying prediction horizons.

\textbf{Dynamic Human Obstacles}
The system integrates both predicted trajectories and historical human presence into the occupancy map through morphological operations and temporal decay mechanisms. Human obstacle integration operates on two levels: immediate presence tracking and predictive obstacle placement.
Current human positions and predicted trajectory points are converted to occupancy map obstacles using morphological dilation with radius $r_{human} = 0.25$ meters. For each human position $p_{human}$ in global map coordinates, the system creates circular obstacle regions:
\begin{equation}
    \mathcal{O}_{human}(x,y) = \{(x,y) : ||[x,y] - p_{human}||_2 \leq r_{human}\}.
\end{equation}

The occupancy map is updated by setting $M_{occ}(x,y) = 1$ for all positions within the dilated regions using a disk-shaped structuring element $S_{disk}$ with radius $r_{pixels} = r_{human} \cdot 100 / res_{map}$.

To prevent stale obstacles from impeding navigation, the system employs a temporal decay mechanism. Each obstacle position is tagged with timestamp $t_{create}$ and removed after decay period $T_{decay} = 5$ timesteps:
\begin{equation}
    M_{occ}(x,y) = \begin{cases}
1 & \text{if } t_{current} - t_{create} \leq T_{decay} \\
0 & \text{if } t_{current} - t_{create} > T_{decay}
\end{cases}.
\end{equation}

Expired obstacles are actively cleared through inverse morphological operations, ensuring previously occupied areas become traversable again. The system maintains separate tracking for active obstacles (recently observed) and expired obstacles (beyond decay time), applying dilation for active regions and explicit clearance for expired regions. This approach balances safety by maintaining obstacles around recent human activity while ensuring long-term navigability by removing outdated restrictions.

\subsection{Path and Action Prediction}

The path and action prediction stage converts the socially-aware occupancy map into executable navigation commands through hierarchical planning. The system employs Fast Marching Method (FMM)~\cite{sethian1996fast} for optimal path generation followed by deterministic action selection.

\textbf{Path Planner} The plath planner operates on the enhanced occupancy map $M_{occ}$ that incorporates static obstacles, dynamic human obstacles, and collision history. The planner first creates a traversability map $T(x,y)$ by applying morphological erosion to account for robot dimensions:
\begin{equation}
    T(x,y) = \neg(\text{binary\_dilation}(M_{occ}, S_{robot})),
\end{equation}

where $S_{robot}$ is a disk-shaped structuring element representing the robot's footprint. The goal region $\mathcal{G}$ is dilated using structuring element $S_{goal}$ to create a multi-goal target, and the path planner algorithm computes the distance field $D(x,y)$ by solving the Eikonal equation $|\nabla D| = 1/T$
 with boundary condition $D(g) = 0$ for all $g \in \mathcal{G}$. The optimal path follows the negative gradient of the distance field, providing the shortest collision-free trajectory to the goal.

\textbf{Action Generation}
From the path solution, the system extracts a short-term goal (STG) point $p_{stg}$ along the optimal path at a fixed lookahead distance. The discrete action $a_t \in \{{move\_forward, turn\_left, turn\_right, stop}\}$
 is determined by comparing the relative angle $\theta_{rel}$ between the robot's current heading $\theta_{robot}$ and the direction to STG:
\begin{equation}
\begin{split}
\theta_{rel} &= (\theta_{robot} - \arctan2(p_{stg}[0] - p_{robot}[0], \\
&\quad p_{stg}[1] - p_{robot}[1])) \mod 360°.
\end{split}
\end{equation}

The action selection follows: $turn\_right\ if\   \theta_{rel} > 15°, turn\_left\ if\ \theta_{rel} < -15°, move\_forward\ if\ |\theta_{rel}| \leq 15°$, and $stop$ if the goal is reached or no valid path exists. This deterministic policy ensures consistent navigation behavior while maintaining computational efficiency for real-time execution.

\section{EXPERIMENT}

\input{tables/main_table}

% In this section, we evaluate the performance of our methodology by comparing it to other Zero-Shot ObjectNav baselines. Additionally, We demonstrate the ability of the three agent helpers to solve three failed cases respectively.

\subsection{Datasets}

We evaluate the \textbf{\textit{\NickName}} framework on the Social-HM3D and Social-MP3D datasets~\cite{gong2025cognition}. These datasets are based on photorealistic HM3D~\cite{ramakrishnan2021habitat} and MP3D~\cite{chang2017matterport3d} scenes and contain simulated environments with real human agents. These datasets contain diverse indoor environments with carefully calibrated human density and incorporate natural motion patterns, where humans alternate between moving and standing still at random walking speeds (0.8-1.2 times the robot speed) and goal-directed trajectories. To demonstrate the zero-shot generalization capability of our method, we perform all evaluations directly on the validation sets of these two datasets without any pre-training or fine-tuning on these environments. Social-HM3D contains 844 scenes and 1087 episodes, while Social-MP3D contains 72 scenes and 317 episodes, comprehensively covering a variety of indoor layouts and human-robot interaction scenarios, enabling a robust evaluation of social navigation performance.

\subsection{Experiment Details}

All experiments are conducted using Habitat 3.0 simulator~\cite{puig2023habitat} on NVIDIA RTX 3090 GPUs. The occupancy map is constructed with a resolution of 5 cm/pixel and covers a 20m × 20m area. For human trajectory prediction, we use a history length of 10 timesteps and predict 10 future steps with updates every 10 timesteps. The human obstacle radius is set to 0.25 meters with a temporal decay period of 5 timesteps to balance safety and navigation efficiency. For orientation-based prediction, we project trajectories 0.5 meters ahead with a prediction width of 0.25 meters. The Fast Marching Method (FMM)~\cite{sethian1996fast} planner uses morphological dilation with a disk-shaped structuring element (radius 3 pixels) to account for robot dimensions. Episodes are terminated after a maximum of 500 timesteps, with success defined as reaching within 0.2 meters of the goal while maintaining at least 0.1 meters distance from humans.

\subsection{Metrics}
Our evaluation metrics build upon existing works~\cite{gong2025cognition} and focus on two principal perspectives: task completion and adherence to social navigation objectives. For task completion, we use Success Rate (SR) and Success weighted by Path Length (SPL). For social norms, we evaluate Human-Robot Collision Rate (H-coll) and Personal Space Compliance (PSC). Considering the human collision radius is 0.3m and the robot is 0.25m, the PSC distance threshold is set to 1.0m. To provide a comprehensive assessment of social navigation performance, we introduce a Final Score metric that combines all evaluation dimensions with weights reflecting their relative importance:
$\textbf{Final Score} = 0.4 \times \text{SR} + 0.2 \times \text{SPL} + 0.2 \times \text{PSC} + 0.2 \times (1 - \text{H-coll})$,
where SR and SPL measure navigation effectiveness, PSC evaluates social compliance, and the inverted H-coll term penalizes human collisions. This weighted combination emphasizes task success while ensuring socially appropriate behavior.

\subsection{Baselines}

We compare our \textbf{\textit{\NickName}} framework against established social navigation methods across different algorithmic paradigms. Rule-based methods include A*~\cite{hart1968formal}, a classical optimal path planner for static environments, and ORCA~\cite{van2011reciprocal}, which incorporates dynamic collision avoidance with oracle access to agent positions and velocities. RL-based methods comprise Proximity-Aware~\cite{cancelli2023exploiting}, which introduces auxiliary tasks modeling human distance and direction to capture proximity dynamics, and Falcon~\cite{gong2025cognition}, the recent state-of-the-art future-oriented framework that integrates explicit human trajectory prediction into reinforcement learning, achieving strong performance through auxiliary prediction tasks during training. All methods use only depth observations as visual input to ensure fair comparison. While RL-based baselines require extensive training (\textbf{\textit{2,396 GPU hours for Falcon}}), our approach operates in a zero-shot manner without any environment-specific training.

\subsection{Compare with SOTA Methods}

Tab.~\ref{tab::main_table} presents the comprehensive evaluation results on both Social-HM3D and Social-MP3D datasets. Our  framework achieves remarkable performance compared to state-of-the-art methods while requiring zero training. On Social-HM3D, our method achieves a Final Score of 61.21, slightly outperforming Falcon (60.39) which requires 2396 GPU hours of training. Notably, our approach significantly reduces human collision rates by over 10\% compared to Falcon (30.36\% vs 41.58\%), demonstrating superior social compliance. On Social-MP3D, \textbf{\textit{\NickName}} achieves a Final Score of 56.58 compared to Falcon's 56.19, while maintaining the lowest collision rate (30.54\% vs 48.53\%). These results demonstrate that our framework can match or exceed the performance of heavily trained RL-based methods without requiring any environment-specific training, offering significant practical advantages for real-world deployment.

% \vspace{-0.15cm}
\subsection{Ablation study}

\input{tables/ablation_study}

\begin{figure*}[!t]
\centering
\includegraphics[width=0.8\linewidth]{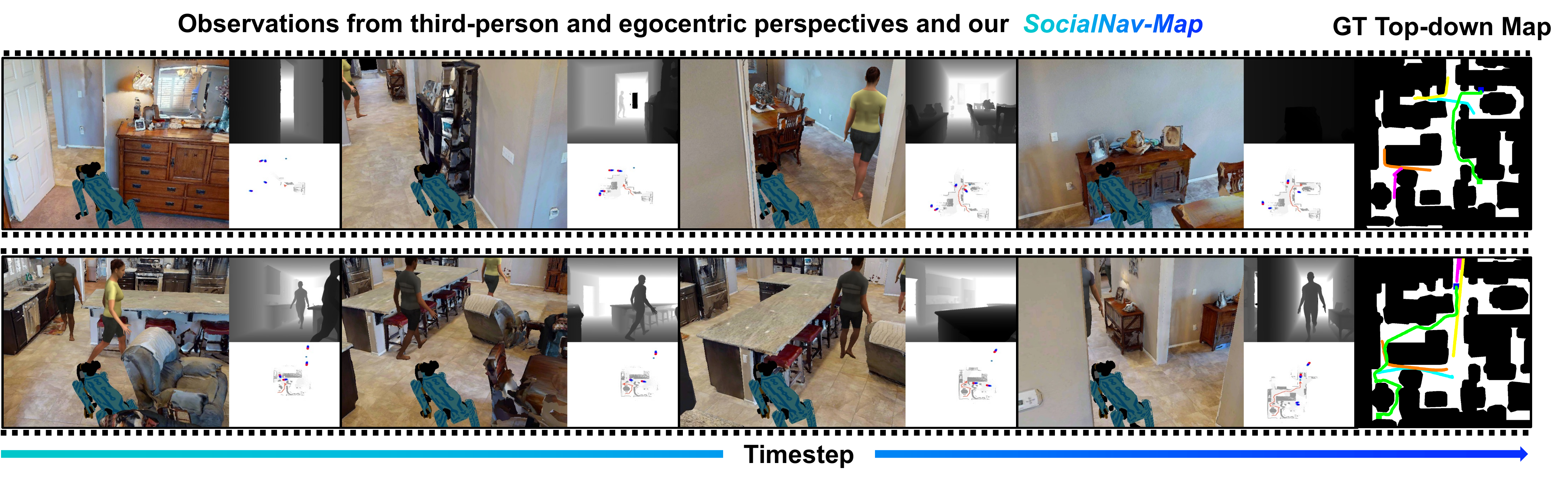}
\vspace{-10pt}
\caption{\textbf{Qualitative results of our \textit{\NickName}.}}
\label{fig5}
\vspace{-10pt}
\end{figure*}

\textbf{Effect of \textit{\NickName}}
We conducted ablation experiments on each component of the \textbf{\textit{\NickName}} across all episodes in the Social-HM3D dataset, demonstrating the efficacy of each component, the results are shown in Tab~\ref{ablation}. 
 Removing the intermediate target mechanism causes the final score to drop from 61.21 to 59.15, while removing history prediction and direction prediction results in scores of 59.82 and 58.94, respectively, demonstrating the critical importance of both trajectory prediction methods. The trajectory fusion component is crucial, with its removal reducing the score to 57.73, highlighting the advantage of combining prediction methods with adaptive weighting. Most importantly, removing dynamic obstacles results in the largest performance drop, to 56.16, confirming that integrating predicted human trajectories as dynamic obstacles is crucial for safe social navigation. These results verify that each component meaningfully contributes to the overall system performance and works together to achieve superior zero-shot social navigation capabilities.

\begin{figure}[!t]
% \vspace{-1cm}
\centering
\includegraphics[width=0.8\linewidth]{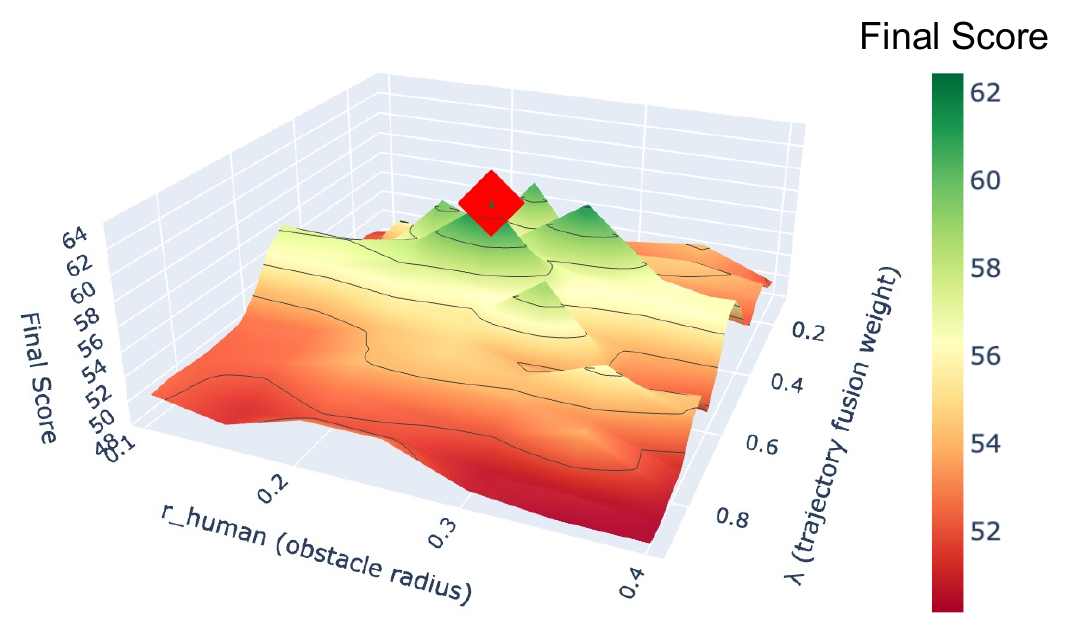}
\caption{\textbf{Hyperparameter Ablation Study on Social-HM3D.} }
\vspace{-0.5cm}
\label{fig4}
\end{figure}

\textbf{Effect of Hyperparameters} 
We analyzed key hyperparameters in our \NickName framework, focusing on the human obstacle radius ($r_{human}$) and trajectory fusion weight ($\alpha$). As shown in Fig.~\ref{fig4}, optimal performance occurs at $r_{human} = 0.25$ meters and $\alpha = 0.5$, achieving a peak Final Score of approximately 61.2. The human obstacle radius defines a safety buffer around predicted human positions; smaller values may lead to unsafe navigation, while larger values result in excessive caution. The trajectory fusion weight $\alpha$ effectively balances history-based and orientation-based predictions, with $\alpha = 0.5$ providing the best integration of both approaches.
% We conducted a detailed analysis of key hyperparameters in our \NickName framework, focusing on the human obstacle radius ($r_{human}$) and trajectory fusion weight ($\alpha$). Our findings, illustrated in Fig.~\ref{fig4}, indicate that optimal performance is achieved with $r_{human} = 0.25$ meters and $\alpha = 0.5$, resulting in a peak Final Score of approximately 61.2. The human obstacle radius defines a safety buffer around predicted human positions; smaller values risk unsafe navigation, while larger values result in overly cautious behavior. The trajectory fusion weight $\alpha$ optimally balances contributions from history-based and orientation-based predictions, with $\alpha = 0.5$ offering the best integration of both methods.

% We conducted a comprehensive analysis of key hyperparameters in our \textbf{\textit{\NickName}} framework, specifically examining the human obstacle radius ($r_{human}$) and trajectory fusion weight ($\alpha$). As shown in Fig.~\ref{fig4}, the performance surface reveals that optimal results are achieved when $r_{human} = 0.25$ meters and $\alpha = 0.5$, yielding the highest Final Score of approximately 61.2. The human obstacle radius represents the safety buffer around predicted human positions, where smaller values may lead to unsafe navigation while larger values create overly conservative behavior. The trajectory fusion weight $\alpha$ balances the contribution between history-based and orientation-based predictions, with $\alpha = 0.5$ providing optimal integration of both methods. 

\subsection{Qualitative Analysis}
Fig.~\ref{fig5} demonstrates the effectiveness of our \textbf{\textit{\NickName}} framework across two navigation episodes. The figure shows the robot successfully avoiding multiple human obstacles while advancing toward its goal. The dynamic occupancy map demonstrates how predicted human trajectories can be integrated with dynamic obstacles to enable proactive path planning. In both episodes, the agent efficiently reached its destination while maintaining a safe distance from humans. The social map continuously updates obstacles and removes outdated predictions to prevent overly conservative behavior. Results demonstrates the effectiveness of our zero-shot social navigation framework.
\renewcommand{\arraystretch}{1.0}
% \vspace{-0.1cm}

\section{CONCLUSIONS}
In this paper, we present \NickName, an innovative zero-shot social navigation framework that requires no training for specific environments while achieving state-of-the-art performance. Our approach combines dynamic occupancy mapping with a dual human trajectory prediction system—history and orientation prediction—to effectively avoid collisions in densely populated settings. Unlike existing reinforcement learning methods that demand 2,396 GPU hours of training, our framework outperforms previous state-of-the-art methods on Social-HM3D and Social-MP3D while reducing human collision rates by over 10\%. By treating predicted human trajectories as dynamic obstacles and employing a temporal decay mechanism, SocialNav-Map excels across multiple datasets without fine-tuning, successfully addressing the generalization limitations of existing approaches and paving the way for practical deployment of social navigation systems in real-world environments with diverse human behaviors.

% We present \textbf{\textit{\NickName}}, a novel zero-shot social navigation framework that requires no training for specific environments while achieving state-of-the-art performance. Our approach combines dynamic occupancy mapping with a dual human trajectory prediction method (history prediction and orientation prediction) to enable active collision avoidance in densely populated environments. Unlike existing reinforcement learning-based methods that require 2,396 GPU hours of training,
% Extensive experiments demonstrate that our zero-shot framework achieves superior performance on Social-HM3D and Social-MP3D, surpassing previous SOTA methods without requiring any training while significantly reducing the human collision rate by over 10\%. By combining predicted human trajectories as dynamic obstacles with a temporal decay mechanism, SocialNav-Map achieves excellent performance on multiple datasets without fine-tuning, addressing key generalization limitations of existing methods and paving the way for practical deployment of social navigation systems in real-world environments with diverse human behaviors.

% \printbibliography

\bibliographystyle{IEEEtran}
\bibliography{main}

\clearpage

\end{document}

%% file: tables/main_table.tex
\begin{table*}[ht]
\centering
\vspace{2pt}
\resizebox{0.88\textwidth}{!}{
\begin{tabular}{cclccccc}
\toprule
\textbf{Dataset} &\multicolumn{2}{c}{\textbf{Method}} & \cellcolor{gray!8}\textbf{SR} $\uparrow$ & \cellcolor{gray!8}\textbf{SPL} $\uparrow$ & \cellcolor{gray!8}\textbf{PSC} $\uparrow$ & \cellcolor{gray!8}\textbf{H-Coll} $\downarrow$ & \cellcolor{gray!8}\textbf{Final Score} $\uparrow$ \\
\midrule
\multirow{5}{*}{Social-HM3D} 
    & \multirow{2}{*}{Rule-Based} 
         & A$^*$~\cite{hart1968formal}                     & { 46.14\scriptsize{$\pm0.7$}} & { 46.14\scriptsize{$\pm0.7$}}         & { 90.56\scriptsize{$\pm0.2$}}           & 53.50\scriptsize{$\pm0.9$} & { { 55.09\scriptsize{$\pm0.7$}}} \\
    &    & ORCA~\cite{van2011reciprocal}                   & 38.91\scriptsize{$\pm0.1$}      &  38.91\scriptsize{$\pm0.1$}              & 90.55\scriptsize{$\pm0.4$}                & 47.52\scriptsize{$\pm1.7$} & { { 51.95\scriptsize{$\pm0.9$}}} \\
    \cmidrule(lr){2-8} 
    & \multirow{2}{*}{RL-Based} 
         & Proximity-Aware~\cite{cancelli2023exploiting}   & 20.11\scriptsize{$\pm1.3$}      & 18.57\scriptsize{$\pm1.9$}                & 92.91\scriptsize{$\pm0.5$}    & 33.99\scriptsize{$\pm0.7$} & { { 43.54\scriptsize{$\pm1.0$}}} \\ 
    &    & { Falcon~\cite{gong2025cognition}}                    & { 53.84}\scriptsize{$\pm 0.6$}  & { 49.30}\scriptsize{$\pm0.7$} &  {89.47}\scriptsize{$\pm1.4$} & {{ 41.58\scriptsize{$\pm1.1$}}} & { { 60.39\scriptsize{$\pm0.8$}}} \\
    \cmidrule(lr){2-8} 
    & \cellcolor{blue!8} \textbf{Zero-Shot}
     & {\cellcolor{blue!8} \textbf{\NickName (Ours)}}                    & {\cellcolor{blue!8} \textbf{51.71}}\scriptsize{$\pm 0.3$}  & {\cellcolor{blue!8} \textbf{43.91}}\scriptsize{$\pm0.5$} & \cellcolor{blue!8} \textbf{{89.12}\scriptsize{$\pm0.7$}} & {\cellcolor{blue!8}\textbf{{ 30.36\scriptsize{$\pm0.9$}}}} & {\cellcolor{blue!8} \textbf{{ 61.21\scriptsize{$\pm0.5$}}}} \\

\midrule
\midrule
   \multirow{5}{*}{Social-MP3D} 
   & \multirow{2}{*}{Rule-Based} 
        & A$^*$~\cite{hart1968formal}                    & { 43.85\scriptsize{$\pm0.3$}}    & { 43.85\scriptsize{$\pm0.3$}}               & 86.74\scriptsize{$\pm3.4$}              & 57.94\scriptsize{$\pm1.5$} & { { 52.07\scriptsize{$\pm0.8$}}} \\
    &   & ORCA~\cite{van2011reciprocal}                  & 40.38\scriptsize{$\pm0.3$}     & 40.38\scriptsize{$\pm0.3$}               & { 91.76\scriptsize{$\pm0.4$}}              & 47.16\scriptsize{$\pm0.2$} & { { 53.15\scriptsize{$\pm0.6$}}} \\
    \cmidrule(lr){2-8} 
    & \multirow{2}{*}{RL-Based} 
        & Proximity-Aware~\cite{cancelli2023exploiting}   & 18.45\scriptsize{$\pm1.4$}     & 17.09\scriptsize{$\pm2.8$}                & 93.37\scriptsize{$\pm0.9$}    & 32.18\scriptsize{$\pm3.3$} & { { 43.04\scriptsize{$\pm0.8$}}} \\
    &    &  {  Falcon~\cite{gong2025cognition}}                    & { 48.47\scriptsize{$\pm0.8$}}  & { 42.08\scriptsize{$\pm0.4$}} &  { 90.48\scriptsize{$\pm0.2$}}      & { { 48.53\scriptsize{$\pm1.2$}}} & { { 56.19\scriptsize{$\pm0.9$}}} \\
    \cmidrule(lr){2-8} 
    & \cellcolor{blue!8} \textbf{Zero-Shot}
     & {\cellcolor{blue!8} \textbf{\NickName (Ours)}}                    & {\cellcolor{blue!8} \textbf{43.53}}\scriptsize{$\pm 0.9$}  & {\cellcolor{blue!8} \textbf{37.09}}\scriptsize{$\pm0.5$} & \cellcolor{blue!8} \textbf{89.30}\scriptsize{$\pm0.2$} & {\cellcolor{blue!8}{ \textbf{30.54\scriptsize{$\pm1.0$}}}} & {\cellcolor{blue!8} \textbf{56.58\scriptsize{$\pm0.6$}}} \\
     
 \bottomrule
\end{tabular}
}
\caption{\textbf{Performance Evaluation of SocialNav Tasks comparing our zero-shot SocialNav-Map framework with Rule-Based and RL-Based Methods on Social-HM3D (upper group) and Social-MP3D (lower group).} Data in the table represents percentages. Falcon~\cite{gong2025cognition} results are reproduced from our implementation.}
\label{tab::main_table}
\vspace{-1em}
\end{table*}

%% file: tables/ablation_study.tex
\begin{table}[!t]
\scriptsize

    \centering
    \renewcommand{\arraystretch}{1.2}
    \setlength{\tabcolsep}{4pt}
    \resizebox{0.88\linewidth}{!}{
    \begin{tabular}{lccccc}
    \toprule[1.0pt]
    
        \textbf{Methods}& \textbf{SR}$\uparrow$ & \textbf{SPL}$\uparrow$ & \textbf{PSC}$\uparrow$ & \textbf{H-Coll}$\downarrow$ &\textbf{Final Score}$\uparrow$\\ \midrule
        % \cmidrule(r){2-2} \cmidrule(lr){3-4}  
    
        \rowcolor{red!6}\textbf{\textbf{\NickName} (Full)} & \textbf{51.71} & \textbf{43.91} & \textbf{89.12} & \textbf{30.36} & \textbf{61.21} \\ \midrule
        w/o Intermediate Goal & 49.24 & 41.18 & 88.95 & 32.47 & 59.15 \\
        w/o History Prediction & 50.13 & 42.26 & 88.74 & 31.89 & 59.82 \\
        w/o Orientation Prediction & 45.87 & 41.95 & 88.46 & 38.12 & 55.94 \\ 
        w/o Trajectory Fusion & 48.69 & 40.81 & 88.32 & 34.58 & 57.73 \\  
        w/o Dynamic Obstacles & 47.35 & 39.67 & 87.29 & 36.94 & 56.16 \\  
    \bottomrule[1.0pt]
    \end{tabular}}
\caption{\textbf{Results of ablation study on Social-HM3D.}}
\label{ablation}
\vspace{-20pt}
\end{table}